\documentclass[fleqn,11pt,oneside]{article}

\usepackage{booktabs} 
\usepackage[pagebackref,breaklinks]{hyperref} 
\usepackage{orcidlink}  
\usepackage{setspace} 
\usepackage{bbm}  
\usepackage{bm}  
\usepackage[normalem]{ulem} 
\usepackage{amsmath,amsthm,amssymb,amscd} 
\usepackage{natbib} 
\usepackage[top=1.1in, bottom=1.1in, left=1.2in, right=1.1in]{geometry} 
\usepackage{caption} 
\usepackage{chngcntr} 

\usepackage{etoolbox}
\newcommand{\insertfig}{\includegraphics[width=\linewidth]{images/teaser_image.pdf}%
\captionof{figure}{Injection loss guidance (iLGD) uses attention injection and
  loss guidance to generate high quality images conforming to a given
  layout.  The first column of each set of images depicts the bounding box
  input to the diffusion model. The second column is the output of Stable
  Diffusion alone. The third column is our method, using the same random
  seeds.\label{fig:teaser}}} 
\makeatletter
\apptocmd{\@maketitle}{\centering\insertfig}{}{}
\makeatother

\newcommand{\etal}{\textit{et al.}\xspace}

\renewcommand{\cite}{\citep}

\newcommand{\td}[1][]{\textcolor{red}{\fcolorbox{red}{pink}{%
\textbf{\textcolor{red}{\scalebox{0.7}[1.0]{\small TODO}}}}%
\ifthenelse{\equal{#1}{}}{}{~\emph{#1}}}\xspace}

\usepackage[ruled]{algorithm2e}

\SetAlFnt{\small}
\SetAlCapFnt{\small}
\SetAlCapNameFnt{\small}
\SetAlCapHSkip{0pt}

\def\R{\mathbbm{R}}
\def\1{\mathbbm{1}}

\newcommand{\bx}{\mathbf{x}}

\newcommand{\argmin}{\mathop{\mathrm{arg\,min}}}

\usepackage{xspace}

\newif\ifrevision \revisiontrue
\newcommand{\revision}[1]{\ifrevision \textcolor{red}{#1}\xspace\else #1\xspace\fi}

\revisionfalse

\usepackage[capitalize]{cleveref}
\crefname{section}{Sec.}{Secs.}
\Crefname{section}{Section}{Sections}
\Crefname{table}{Table}{Tables}
\crefname{table}{Tab.}{Tabs.}

\begin{document}
\title{\vspace*{-6ex}Enhancing Image Layout Control with Loss-Guided Diffusion Models\vspace*{-1ex}}
\author{Zakaria Patel\thanks{\href{https://www.ecomtent.ai/}{Ecomtent} \& Department of Computer Science,
University of Toronto \newline\hspace*{1.8em}Corresponding author. Email:
\href{mailto:zakaria.patel@mail.utoronto.ca}{zakaria.patel@mail.utoronto.ca}}
\,\orcidlink{0009-0003-1752-1910} 
\and Kirill Serkh
\thanks{Departments of Mathematics and Computer Science,
University of Toronto\newline\hspace*{1.8em}Email:
\href{mailto:kserkh@math.toronto.edu}{kserkh@math.toronto.edu}}
\,\orcidlink{0000-0003-4751-305X}}

\date{\today\vspace*{-4ex}}


\newgeometry{bottom=0.6in}
\maketitle
\thispagestyle{empty}

\vspace*{-2ex}
\begin{abstract}
\small
    Diffusion models are a powerful class of generative models capable of
    producing high-quality images from pure noise using a simple text
    prompt.  While most methods which introduce additional spatial
    constraints into the generated images (e.g., bounding boxes) require
    fine-tuning, a smaller and more recent subset of these methods take
    advantage of the models' attention mechanism, and are training-free.
    These methods generally fall into one of two categories.  The first
    entails modifying the cross-attention maps of specific tokens directly
    to enhance the signal in certain regions of the image. The second works
    by defining a loss function over the cross-attention maps, and using the
    gradient of this loss to guide the latent. While previous work explores
    these as alternative strategies, we provide an interpretation for these
    methods which highlights their complimentary features, and demonstrate
    that it is possible to obtain superior performance when both methods are
    used in concert.
\end{abstract}

\restoregeometry


\section{Introduction}
\label{sec:intro}

Recently, diffusion models~\cite{sohl2015deep, ho2020denoising,
song2020generative} have emerged as a powerful class of generative models
capable of producing high quality samples with superior mode coverage. These
models are trained to approximate a data distribution, and sampling from
this learned distribution can subsequently generate very realistic and
diverse images. Incorporating conditioning~\cite{saharia2022photorealistic,
ramesh2022hierarchical, rombach2022high} further extends the utility of
diffusion models by allowing one to specify the contents of the desired
image using a simple text prompt, leveraging the models' impressive
compositional capabilities to combine concepts in ways that may not have been present in the training set. Conditioning on a text prompt does
not, however, allow one fine-grained control over the layout of the final
image, which is instead highly dependent on the initial noise sample. 

An especially simple and intuitive way of describing a layout is to provide
bounding boxes for various tokens in the text prompt.
One way to realize a model taking such an input is to resort to
training-based methods, wherein a pretrained model undergoes additional
finetuning using training data where the images have been supplemented by
their layouts.
While such methods can achieve impressive performance~\cite{zheng2023layoutdiffusion, zhang2023adding},
this often involves the
introduction of additional model complexity and training cost, in addition
to the difficult task of compiling the training data. 
A recently
proposed alternative approach~\cite{xie2023boxdiff, chen2023trainingfree}
uses the cross-attention module to achieve training-free layout control.

%

In this paper, we propose injection loss guidance (iLGD), a training-free
framework in
which the model's denoising process works in synergy with loss guidance,
such that sampled images simultaneously present the appropriate layout and
maintain good image quality. First, we bias the latent towards the desired
layout by altering the model's attention maps directly, a process which we
refer to as attention injection. The goal of injection is not to produce a
latent which perfectly adheres to the desired layout, but rather to provide
a comfortable starting point for loss guidance. The resulting latent is
sufficiently close to the desired layout that we can afford to use a smaller amount of loss guidance. In doing so, we reimagine the role of injection
as a coarse biasing of the diffusion process, and loss guidance as a
refiner. We show that such a framework is capable of controlling the layout,
while achieving superior image quality in many cases. 


\section{Related Work}
\label{sec:related-work}

\subsection{Generative Models}

Generative models learn to estimate a data distribution with the goal of
generating samples from this distribution. 
%
%
\revision{Recently}, a new family of generative models, known as diffusion models~\cite{ sohl2015deep, ho2020denoising, song2020generative}, have achieved
superior results on image synthesis compared to the previous state of the
art~\cite{dhariwal2021diffusion}, with samples exhibiting incredible
diversity and image quality. While early diffusion models were formulated as
Markov chains and relied on a large number of transitions to generate
samples, Song \etal~\cite{song2020score} showed that such a sampling procedure can be
viewed as a discretization of a certain stochastic differential equation
(SDE). In particular, Song \etal~\cite{song2020score} showed that there exists a
family of SDEs, whose solutions are sampling trajectories from the diffusion
model. One such SDE,  called the probability flow ODE, is completely
deterministic and contains no noise. This enables the use of various 
ordinary differential equation (ODE) solvers for efficient sampling~\cite{karras2022elucidating, lu2022dpm}.

\subsection{Controllable Generation}

Diffusion models can use a wide range of techniques for controllable
generation. SDEdit~\cite{meng2022sdedit} allows the user to specify a layout
by using paint strokes, which are noised to time $t < T$ to provide an
initialization for solving the reverse-time SDE.  The realism of the final
image is, however, sensitive to the initial noise level~$\sigma_t$, and
guiding the generation of new images requires all high-level features to be
specified in the stroke image. A more user-friendly method by Voynov \etal~\cite{voynov2023sketch} requires only a simple sketch to guide the
denoising process. The user-provided sketch is compared with the edges
extracted by a latent edge predictor in order to compute a loss, which is
used to iteratively refine the latent. In this case, the latent edge
predictor must be trained, and sketches of more complicated scenes may be
tedious. Zhang \etal~\cite{zhang2023adding} propose a more general method in which a
separate encoder network takes as input a conditioning control image, such
as a sketch, depth map, or scribble, to guide the generation process.
Unfortunately, this requires finetuning a large pretrained encoder. When the
inputs are specified as bounding boxes, smaller trainable modules can be
used between the layers of the denoising UNet to encode layout
information~\cite{zheng2023layoutdiffusion}. Cheng \etal~\cite{cheng2023layoutdiffuse}
also use an additional module which takes in bounding box inputs, injecting
it directly after the self-attention layers. Once again, however, neither
method can be adapted immediately, as the additional parameters necessitate
further training.

A number of other methods have been proposed that are training-free. Bansal \etal~\cite{bansal2024universal} define a generic loss on the noiseless latent
$\hat{\mathbf{z}}_0$ predicted from $\mathbf{z}_t$ by Tweedie's formula, and
subsequently perform loss guidance on the latent $\mathbf{z}_t$ at each time
step. One downside of this method is that, while the loss expects clean
images, the predicted latent $\hat{\mathbf{z}}_0$ is an approximation to
only an average of possible generated images, and so can be blurry for large
times $t$. An alternative approach \revision{called MultiDiffusion} was proposed by
Bar-Tal \etal~\cite{bar2023multidiffusion}, who used separate score
functions on various regions in a latent diffusion model, with an
optimization step at each iteration designed to fuse the separate diffusion
paths. While this method can be effective in many cases, it can nonetheless
exhibit patchwork artifacts, where the final image appears to be composed of
several images rather than depicting a single scene.

Alternatively, several works have explored the use of cross-attention to
achieve training-free layout control. Hertz \etal~\cite{hertz2023prompttoprompt}
demonstrate how attention maps can be injected from one diffusion process to
another, and reweighted to control the influence of specific tokens.
Subsequent work by Balaji \etal~\cite{balaji2023ediffi} builds upon this idea by
directly manipulating the values in the attention map to obtain the desired
layout, although it is difficult to precisely localize objects appearing in
an image with this method alone. Singh \etal~\cite{singh2022highfidelity} also use
this technique to improve semantic control in stroke-guided image synthesis,
which they combine with loss guidance based on the stroke image to improve
the realism of generated images. Instead of using strokes, Chen \etal~\cite{chen2023trainingfree}
show that controlling layout with bounding boxes is possible by using loss
guidance, where the loss is defined on the attention maps, although this method
requires searching for a suitable noise initialization. Concurrent work
by Xie \etal~\cite{xie2023boxdiff} and Couairon \etal~\cite{couairon2023zero} also use
attention-based loss guidance; the former adds spatial constraints to
control the scale of the generated content, while the latter uses
segmentation maps instead of bounding boxes. Epstein \etal~\cite{epstein2023diffusion}
show that it is even possible to control properties of objects in an image, such
as their shape, size, and appearance, through their attention maps, and
subsequently manipulate these properties through loss guidance. 

These works demonstrate the utility of injection, loss guidance, and the
general role of cross-attention in layout control. We take a joint approach
where we use cross-attention injection to assist loss guidance in producing
the desired layout from simple bounding box inputs. In analyzing the role of each technique in layout
control, we offer justification for their complementary use. The result is a
powerful and intuitive method for layout control which maintains the quality
of the generated images. 


\section{Preliminaries}
\label{sec:preliminaries}

\subsection{Cross-Attention}
To perform conditional image synthesis with text, Stable Diffusion leverages
a cross-attention mechanism~\cite{vaswani2023attention}. Cross-attention enables the modelling of complex
dependencies between two sequences
$\mathbf{X}^T=(\mathbf{x}_1, \mathbf{x}_2, \ldots, \mathbf{x}_n)$ and
$\mathbf{Y}^T=(\mathbf{y}_1, \mathbf{y}_2, \ldots, \mathbf{y}_k)$, whose
elements are projected to query, key and value vectors using projection
matrices
\begin{align*}
&\mathbf{X}\mathbf{W_q} = \mathbf{Q} \in \R^{n \times d_k},\\
&\mathbf{Y}\mathbf{W_k} = \mathbf{K} \in \R^{k \times d_k},\\
&\mathbf{Y}\mathbf{W_v} = \mathbf{V} \in \R^{k \times d_v},
\end{align*}
\revision{where $d_k$ denotes the dimension of the query and key vectors,
and $d_v$ denotes the dimension of the value vectors.}  Subsequently, the
attention weights \revision{are} computed as
\begin{align*}
    A=\text{Softmax}\left(\frac{\mathbf{QK}^T}{\sqrt{d_k}}\right) \in
    \R^{n\times k}.
\end{align*}
The new representation for the sequence $\mathbf{X}$ is
\begin{align*}
\mathbf{Z} = A\mathbf{V} \in \R^{n \times d_v}.
\end{align*}

In diffusion models, the sequence $\mathbf{X}$ represents the image, where
each $\mathbf{x}_i$ represents a pixel, and $\mathbf{Y}$ is a sequence of
token embeddings. The attention weights $A$, also called the attention or
cross-attention map, follow the same spatial arrangement as the image, and a
unique map $A_j$ is produced for each token $\mathbf{y}_j$ in
$\mathbf{Y}$. Each entry $A_{ij}$ describes how strongly related a
spatial location $\mathbf{x}_i$ is to the token $\mathbf{y}_j$. We leverage
this feature of cross-attention to guide the image generation process. 

\subsection{Score Matching}
The forward and reverse processes can be modelled by solutions of stochastic
differential equations (SDEs)~\cite{song2020score}. While determining
the coefficients of the forward process SDE
is straightforward, the reverse process corresponds to a solution of the
reverse-time SDE, which requires learning the score of the intractable
marginal distribution $q(\mathbf{x}_t)$. Instead, Song \etal~\cite{song2020score}
use the score-matching objective 
\begin{equation}
  \mathbbm{E}_t [ \lambda(t)
  \mathbbm{E}_{q(\mathbf{x}_0)}\mathbbm{E}_{q(\mathbf{x}_t|\mathbf{x}_0)} [
  \lVert \mathbf{s_\theta}(\mathbf{x}_t, t) - \nabla_{\mathbf{x}_t} \log
  q(\mathbf{x}_t|\mathbf{x}_0) \rVert^2_2 ] ],
    \label{eq:score_matching}
\end{equation}
for some \revision{positive} function $\lambda\colon [0,T] \to \R$.

While Eq.~\eqref{eq:score_matching} does not directly enforce learning the
score of
$q(\mathbf{x}_t)$, it is nonetheless minimized when
$\mathbf{s}_\theta(\mathbf{x}_t,t) =\nabla_{\mathbf{x}_t} \log
q(\mathbf{x_t})$~\cite{vincent2011connection}.
Conditioning on
$\mathbf{x}_0$ provides a tractable way to obtain a neural network
$\mathbf{s}_\theta(\mathbf{x}_t, t)$ which, given enough parameters, matches
$\nabla_{\mathbf{x}_t} \log q(\mathbf{x}_t)$ almost everywhere. Because
the forward process $q(\mathbf{x}_t|\mathbf{x}_0)$ is available in
closed form, it can be shown that the neural network which minimizes
this loss is 
\begin{equation}
\label{eq:predicted_score}
    \mathbf{s}_\theta(\mathbf{x}_t, t) =
    -\frac{\bm{\epsilon}_\theta(\mathbf{x}_t,
    t)}{\sigma_t},
\end{equation}
where $\sigma_t$ is the standard deviation of the forward process at time
$t$, and $\bm{\epsilon}_\theta(\mathbf{x}_t, t)$ predicts the scaled noise
$\bm{\epsilon} \sim \mathcal{N}(\mathbf{0},\mathbf{I})$ in $\mathbf{x}_t$.

When the predicted score function is available, any one of a family of reverse-time
SDEs, all with the same marginal distributions $p_\theta(\mathbf{x}_t) \approx
q(\mathbf{x}_t)$, can be solved to sample from $p_\theta(\mathbf{x}_0)
\approx q(\mathbf{x}_0)$. One of these SDEs is noise-free, and is known as
the probability flow ODE. A highly efficient way to sample from
$p_\theta(\mathbf{x}_0)$ is to solve the probability flow ODE using a small
number of large timesteps~\cite{song2020score}. An important benefit of sampling using 
an ODE is that the sampling process is deterministic, in the sense that it associates
each noisy image $\mathbf{x}_T$ with a unique noise-free sample
$\mathbf{x}_0$~\cite{song2020score}.

\revision{
In practice, the denoising process usually operates on a latent variable
$\mathbf{z}_t$, where an autoencoder is used to project to and from image
space $\mathbf{x}_t$.  In order to generate images which correspond to
a user-supplied text prompt, the noise predictor
$\bm\epsilon_\theta(\mathbf{z}_t,t,\mathbf{y})$ is conditioned on text
prompts $\mathbf{y}$, and samples are computed using this noise predictor by
classifier-free guidance (CFG).  In this paper, we denote the CFG noise
prediction by $\bm{\epsilon}_\theta(\mathbf{z}_t,t,\mathbf{y})$, or
sometimes just $\bm{\epsilon}_\theta(\mathbf{z}_t,t)$, when the dependence
on $\mathbf{y}$ is clear. We provide more details about diffusion models in
\autoref{app:diffusion}.  }

\subsection{Controllable Layout Generation}
We use BoxDiff~\cite{xie2023boxdiff}\revision{, the method of Chen
\etal~\cite{chen2023trainingfree}, and
MultiDiffusion~\cite{bar2023multidiffusion} as our primary points} of
comparison. In~\cite{xie2023boxdiff} and~\cite{chen2023trainingfree},
the authors apply spatial constraints on the attention
maps of a latent diffusion model to derive a loss, and directly update the
latent at time step $t$ by replacing it with
\begin{equation}
    \label{eq:boxdiff}
    \mathbf{z}'_t = \mathbf{z}_t - \alpha_t \cdot
    \nabla_{\mathbf{z}_t} \mathcal{L}_\mathbf{y}(\mathbf{z}_t),
\end{equation}
where $\mathcal{L}_\mathbf{y}$ is a loss function which depends on the neural
network, the prompt $\mathbf{y}$, and
a set of bounding boxes.
\revision{
The parameter $\alpha_t$ controls the strength of the loss guidance in each
iteration.  In BoxDiff, $\alpha_t$ decays linearly from $t=T$ to $t=0$,
while in Chen \etal, $\alpha_t = \eta\cdot\sigma_t^2$ for some scale factor
$\eta > 0$.
In MultiDiffusion, letting $\Phi_\theta(\mathbf{z}_t,t,\mathbf{y})$ denote
the function which computes $\mathbf{z}_{t-1}$ from $\mathbf{z}_t$ 
using the predicted scaled
noise ${\bm\epsilon}_\theta(\mathbf{z}_t,t,\mathbf{y})$, a new latent
$\mathbf{z}'_{t-1}$ is produced by solving the optimization problem
\begin{align*}
\mathbf{z}'_{t-1} = \argmin_\mathbf{z} \sum_{i=1}^k \bigl\| \mathbf{m}_i \otimes
\bigl(\mathbf{z} - \Phi_\theta(\mathbf{z}_t,t,\mathbf{y}_i)\bigr) \bigr\|^2,
\end{align*}
where each $\mathbf{m}_i\in \{0,1\}^{n}$ is a mask
corresponding to the bounding box of the prompt $\mathbf{y}_i$.
}




\section{Method}

\begin{figure*}
    \centering
    \includegraphics[width=0.8\textwidth]{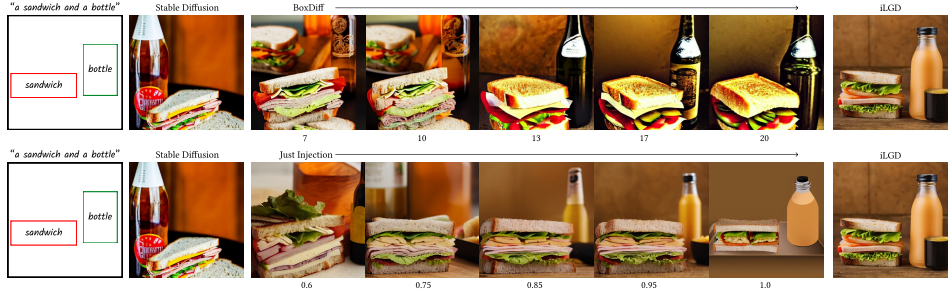}
    \caption{The effects of varying the strengths of BoxDiff and attention injection
    by tuning
    their respective parameters. 
    In the top row, we sweep through various choices of $\alpha_T$ to tune the
    guidance strength of BoxDiff. In the bottom row,
    we sweep through various choices of the injection strength $\nu'$.
    For iLGD, shown in the final column, we use $\nu'=0.75$ and $\eta =
    0.8$.}
    \label{fig:inject_and_boxdiff_sweep}
\end{figure*}

\subsection{Attention Injection}

Hertz \etal~\cite{hertz2023prompttoprompt} observed that, by extracting the attention
maps from a latent diffusion process and applying them to another
one with a modified token sequence, it is possible to transfer the composition
of an image. This technique is very effective, but requires an original set
of attention maps which produce the desired layout, and it is not feasible
to generate images until this layout is obtained.

Instead, we rely on the observation that the attention maps early in the
diffusion process are strong indicators of the generated image's
composition.  For early timesteps, both the latents and the attention maps
are relatively diffuse, and don't suggest any fine details about objects in
the image.  Motivated by this, we manipulate the attention maps by
artificially enhancing the signal in certain regions.  Given a list of
target tokens $S \subset \{1,2,\ldots,k\}$, we define $\mathbf{m} = \{ \mathbf{m}_1,
\mathbf{m}_2,\ldots, \mathbf{m}_k\}\in \R^{n\times k}$ by setting each mask
$\mathbf{m}_j$, $j \in S$, equal to $1$ over the region that the text token
$\mathbf{y}_j$ should correspond to, and zero otherwise, and perform
injection by replacing the cross-attention map $A_t$ at time $t$ with
\begin{equation}
    \label{eq:injection}
    A'_t = \text{softmax}\left(\frac{\mathbf{Q}_t\mathbf{K}_t^T + \nu_t
    \mathbf{m}}{\sqrt{d_k}}\right),
\end{equation}
\revision{where} $\nu_t>0$. We follow Balaji \etal~\cite{balaji2023ediffi}
and use the scaling
\begin{equation}
    \label{eq:injection_strength}
    \nu_t = \nu' \cdot \log(1 + \sigma_t) \cdot \max(\mathbf{Q}_t\mathbf{K}_t^{T}).
\end{equation}
Scaling by $\sigma_t$ ensures the injection strength is appropriate for a
given timestep, and $\nu'$ is a constant which controls the overall strength of
injection. 

This way, we directly bias the model's predicted score
$\mathbf{s}_\theta(\mathbf{x}_t,t) \approx \nabla_{\mathbf{z}_t} \log
{q_t(\mathbf{z}_t|\mathbf{y})}$ so that each latent $\mathbf{z}_{t-1}$
more closely corresponds to the desired layout. We denote the corresponding
modified noise prediction by ${\bm \epsilon}_\theta(\mathbf{z}_t,t,A_t
\xrightarrow{\mbox{\raisebox{-0.4ex}[0pt][0pt]{\tiny$\nu'$,$\mathbf{m}$}}} A_t')$.

\subsection{Loss Guidance}

Conditional latent diffusion models predict the time-dependent conditional
score $\nabla_{\mathbf{z}_t} \log {q(\mathbf{z}_t|\mathbf{y})}$, so that the
resulting latent at the end of the denoising process
is sampled from $q(\mathbf{z}_0|\mathbf{y})$. 
We can modify the conditional score at time
$t$ by introducing a loss term $\ell_\mathbf{y}(\mathbf{z}_t)$,
\begin{equation}
    \label{eq:modified_score}
    \nabla_{\mathbf{z}_t} \log \hat{q}(\mathbf{z}_t|\mathbf{y}) =
    \nabla_{\mathbf{z}_t} \log q(\mathbf{z}_t|\mathbf{y}) - \eta
    \nabla_{\mathbf{z}_t} \ell_\mathbf{y}(\mathbf{z}_t),
\end{equation}
which corresponds to the marginal distribution
\begin{align}
\hat q(\mathbf{z}_t | \mathbf{y}) \propto q(\mathbf{z}_t | \mathbf{y})
e^{-\eta \ell_\mathbf{y}(\mathbf{z}_t)}.
\end{align}
The scaling constant $\eta$ controls the relative strength of loss guidance.
By using annealed Langevin dynamics~\cite{song2020generative}, it is
possible to use the predicted score function together with the loss term to
sample from an
approximation to $\hat q(\mathbf{z}_0 | \mathbf{y})$. While this provides a clear
interpretation for the effect of loss guidance, the cost of annealed Langevin
dynamics can be fairly large. Instead, we solve the probability flow ODE using the
score function 
\begin{align}
\hat{\mathbf{s}}_\theta(\mathbf{z}_t,t) := 
\mathbf{s}_\theta(\mathbf{z}_t,t) - \eta \nabla_{\mathbf{z}_t}
\ell_\mathbf{y}(\mathbf{z}_t).
\end{align}
This no longer corresponds to sampling from an approximation to $\hat
q(\mathbf{z}_0 | \mathbf{y})$ (see~\cite{song2023loss}), however, so long as
the latents $\mathbf{z}_t$ are not too out-of-distribution with respect to
the marginals $q(\mathbf{z}_t|\mathbf{y})$, then
this process influences the trajectory to favor samples from
$p_\theta(\mathbf{z}_0| \mathbf{y})$ for which the loss term is small.
%
%

For layout control, given a list of target tokens $S \subset
\{1,2,\ldots,k\}$, we choose the simple loss function
\begin{equation}
    \label{eq:lg_loss_term}
    \ell_\mathbf{y}(\mathbf{z}_t) = \sum_{j\in S}
    \revision{\text{sum}}(\mathbf{\bar m}_j \odot (A_t)_j) -
    \revision{\text{sum}}(\mathbf{m}_j \odot (A_t)_j),
\end{equation}
where $\mathbf{m}_j$ is a mask whose value is $1$ over the region where
token $\mathbf{y}_j$ should appear, and $0$ otherwise, and
$\mathbf{\bar m}_j= 1-\mathbf{m}_j$. Intuitively, this simple loss
encourages \revision{the} sampling \revision{of} latents whose attention
maps for each target token take their largest values within the masked
regions.

\revision{When high levels of loss-guidance are used, the specific choice
of the loss function heavily influences the behavior of the denoising
process.  In BoxDiff, Xie \etal~\cite{xie2023boxdiff} use
sums over the $P \ll n$  most attended-to pixels in the masked attention maps,
while in Chen \etal~\cite{chen2023trainingfree} the authors use a
normalized loss which depends on the ratio $\text{sum}(\mathbf{m}_j
\odot (A_t)_j)/\text{sum}((A_t)_j)$.  These choices seem to be essential for
enabling those methods to maintain good image quality at very high levels of
loss guidance. Since we will use much lower levels of loss guidance, we find
that our method is significantly less sensitive to the choice of loss
function, with the result that our simple loss function works well to contain the
attentions within the regions defined by $\mathbf{m}$. }

In practice, diffusion models are trained to predict the scaled noise
$\bm\epsilon \sim \mathcal{N}(\mathbf{0},\mathbf{I})$ in $\mathbf{z}_t$.
Revisiting Eq.~\eqref{eq:predicted_score}, we observe that we can define the
corresponding modified noise prediction by scaling the loss-guidance term
appropriately~\cite{dhariwal2021diffusion}:
\begin{equation}
    \hat{\bm{\epsilon}}_\theta(\mathbf{z}_t, t) :=
    \bm{\epsilon}_\theta(\mathbf{z}_t, t) +
    \eta\sigma_t
    \nabla_{\mathbf{z}_t}\ell_\mathbf{y}(\mathbf{z}_t). 
\end{equation}

\revision{
Since the loss function $\ell_\mathbf{y}(\mathbf{z}_t)$ is a scalar function,
computing its gradient with respect to $\mathbf{z}_t$
by reverse accumulation requires only a single 
sweep through the computational graph. As a result, the runtime and memory
requirements of sampling using loss guidance are approximately twice
those of sampling without loss guidance. This can be seen in, for example,
Table~1 of Chen \etal~\cite{chen2023trainingfree}.}

\subsection{iLGD}

\begin{figure*}
    \centering
    \includegraphics[width=0.8\textwidth]{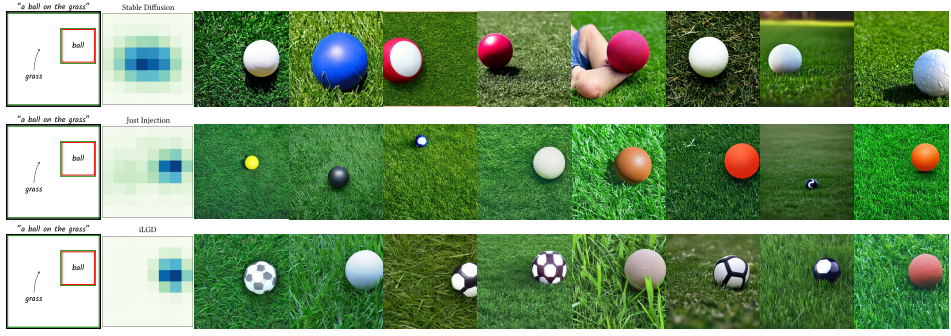}
    \caption{Images generated with the prompt ``a ball on the grass,'' using
    the bounding boxes shown in the first column. Each row corresponds to a
    different method. The bounding boxes in the first column are used for injection
    and iLGD. The attention maps in the second column are averages over the
    $8\times8$ resolution attention maps at $t=0$ over 100 random seeds. Each of the
    8 columns of images in this figure corresponds to one of these 100
    seeds.}
    \label{fig:attend_bias}
\end{figure*}

One weakness of attention injection is that, by directly modifying the
attention maps, we disrupt the agreement between the predicted score
function $\mathbf{s}_\theta(\mathbf{z}_t,t)$ and the true conditional score
$\nabla_{\mathbf{z}_t}\log q(\mathbf{z}_t|\mathbf{y})$. The discrepancy
between the actual and predicted score functions becomes especially
pronounced at smaller noise levels, where fine details in the cross
attention maps are destroyed by the injection process. This results in a
cartoon-like appearance in the final sampled images, as seen in Figure~17
of~\cite{balaji2023ediffi} and in the bottom row of
\autoref{fig:inject_and_boxdiff_sweep}. The sensitivity of the
cross-attention maps at low noise levels makes it difficult to use injection to
precisely control the final layout, without negatively
influencing image quality. 
One big advantage of attention injection, however, is that at high noise
levels it is possible to strongly influence the cross-attention maps while
still producing latents that are in-distribution.
Interestingly, even the cartoon-like images resulting from excessive
attention injection at low noise levels are not actually
out-of-distribution, but are only in an unwanted style.

A weakness of loss-guidance is that the ad-hoc choice of the loss function
may not compete well with the predicted score
$\mathbf{s}_{\theta}(\mathbf{z}_t, t)$. In each denoising step, the latent
$\mathbf{z}_{t-1}$ must fall in a high probability region of of
$q(\mathbf{z}_{t-1}|\mathbf{z}_t)$, otherwise the predicted latent
$\mathbf{z}_{t-1}$ will be out-of-distribution, which results in
degraded image quality in the final sampled latent $\mathbf{z}_{0}$. This
means that the guidance term's influence in Eq.~\eqref{eq:modified_score} should
be small enough to avoid moving the latents into low probability regions. On
the other hand, small loss guidance strengths may exert too little influence
on the sampling trajectory. In this case, the model produces in-distribution
samples, but ones which do not fully agree with the desired layout. This
tradeoff is illustrated in the first row of
\autoref{fig:inject_and_boxdiff_sweep}\revision{, which shows images with
unnaturally high contrast and saturation appearing at high loss-guidance
strengths}. One advantage of loss-guidance is that, even at low strengths,
it is able to exert some influence over sampling trajectories without
biasing the style of the images.

We take the point of view that injection is better suited as a course
control over the predicted latents, and cannot fully replace loss guidance.
Instead, we propose using injection together with loss guidance in a
complimentary fashion, in such a way that they compensate for each other's
weaknesses.  We call this approach to layout control injection loss guidance
(iLGD).
Instead of delegating the layout generation task entirely to
loss guidance, we rely on injection to first bias the latent, as illustrated
in the second row of~\autoref{fig:attend_bias}. 
The first row of this figure shows that, when using Stable Diffusion alone,
the ball appears in random locations near the center of the image, while the
second row shows that injection encourages it to appear more frequently
inside of the bounding box, towards the top right.
When we also perform loss guidance, the third row of the figure shows that
the averaged attention map is significantly more concentrated inside of the
bounding box. Since the original score function is modified additively,
the amount of loss guidance can be made sufficiently small so that the
latents stay in-distribution, while still being large enough to
influence the sampling trajectory. This is true even at small noise
levels, when fine details are present in the images.
%
%

We also observe that, when using both injection and loss guidance, the
details of the objects better reflect the context of the scene due to higher
levels of attention on those objects. In the third row, many of the balls
appear in the style of a soccer ball, which is likely the most common type
of ball to appear together with grass.

\begin{figure}
    \centering
    \includegraphics[width=0.45\linewidth]{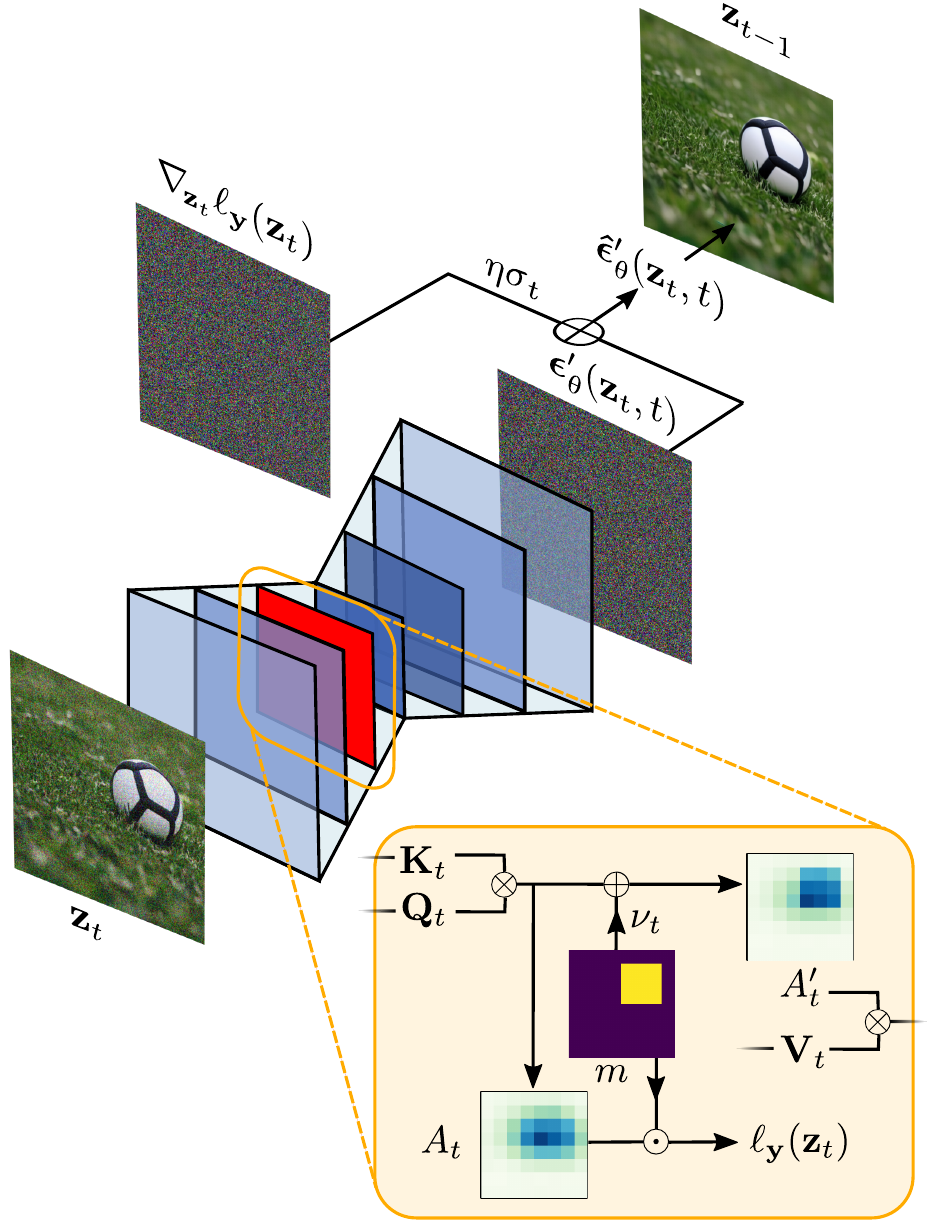}
    \caption{A graphical depiction of injection loss guidance (iLGD).}
    \label{fig:ilgd_algorithm}
\end{figure}

We outline our algorithm in \autoref{alg:ilgd} and depict it
visually in \autoref{fig:ilgd_algorithm}.  We perform attention
injection from timestep $T$ to $t_{\text{inject}}$ in order to obtain the
modified predicted noise $\bm{\epsilon}'_{\theta}(\mathbf{z}_t, t)$. In each
timestep, we further refine this latent by performing loss-guidance
simultaneously, from timestep $T$ to $t_{\text{loss}}$. In practice, we find
it useful to perform loss-guidance for several more steps after we stop
injection, $t_{\text{loss}} > t_{\text{inject}}$. 

\begin{algorithm}[t]
    \footnotesize
    \SetAlgoNoLine
    \KwIn{A prompt $\mathbf{y}$; a list of target tokens $S \subset
    \{1,2,\ldots,k\}$; a collection of bounding boxes, one for each token in
    $S$; an injection strength $\nu'$; a loss guidance strength $\eta$.}
    \KwOut{The generated image $\mathbf{x}_0 = \mathcal{D}(\mathbf{z}_0)$.}
    Construct the mask $\mathbf{m}=\{
    \mathbf{m}_1,\mathbf{m}_2,\ldots,\mathbf{m}_k\}$ from the bounding
    boxes\;
    Initialize $\mathbf{z}_T \sim \mathcal{N}(\mathbf{0}, \mathbf{I})$\;
    \For{$t = T,\ldots,1$}{
      $\bm\epsilon'_{\theta}(\mathbf{z}_t, t) = \begin{cases}
        \bm\epsilon_\theta(\mathbf{z}_t, t, A_t 
        \xrightarrow{\mbox{\raisebox{-0.4ex}[0pt][0pt]{\tiny$\nu'$,$\mathbf{m}$}}}
        A_t') & \text{if $t > t_\text{inject}$}, \\
        \bm\epsilon_\theta(\mathbf{z}_t, t) & \text{otherwise}
      \end{cases}$\;
      $\ell_{\mathbf{y}}(\mathbf{z}_t) = \begin{cases}
        \sum_{j\in S}
          \revision{\text{sum}}(\mathbf{\bar m}_j \odot (A_t)_j) -
          \revision{\text{sum}}(\mathbf{m}_j \odot (A_t)_j) &
        \text{if $t > t_\text{loss}$}, \\
        0 & \text{otherwise}
      \end{cases}$\;
      $\hat{\bm\epsilon}_\theta'(\mathbf{z}_t, t) = \bm\epsilon'_{\theta}(\mathbf{z}_t, t)
      + \eta \sigma_t \nabla_{\mathbf{z}_t}\ell_{\mathbf{y}}(z_t)$\;
      \text{Compute $\mathbf{z}_{t-1}$ from $\mathbf{z}_t$ using
      $\hat{\bm\epsilon}_\theta'(\mathbf{z}_t, t)$}\;
    }
    \caption{Pseudocode for iLGD}
    \label{alg:ilgd}
\end{algorithm}


\section{Experiments}
\label{sec:experiments}

\subsection{Experimental Setup}

\paragraph{Datasets}
We follow Xie \etal~\cite{xie2023boxdiff} and evaluate performance
on a dataset consisting of 200 prompt and bounding-box pairs,
spanning 20 different prompts and 27 object categories. Each prompt
reflects either of the following prompt structures: ``a \{\} \ldots,'' or
``a \{\} and a \{\} \ldots.'' 
\revision{MultiDiffusion requires separate prompts for the foreground
elements and for the background, which we manually created from our original
prompts. For example, the prompt ``a red book and a clock'' became the two
foreground prompts ``a red book'' and ``a clock,'' with a null background
prompt,  while ``a giraffe on a field'' became the foreground prompt 
``a giraffe'' in the foreground
and the background prompt ``a field.''}

\revision{
\paragraph{Methods}
In this section, we compare our proposed method (iLGD),
BoxDiff~\cite{xie2023boxdiff}, Chen \etal~\cite{chen2023trainingfree},
MultiDiffusion~\cite{bar2023multidiffusion}, and Stable
Diffusion~\cite{rombach2022high}. We also include the results of an ablation
study of our method, in which we use just injection (SD + I) or just loss
guidance (SD + LG) alone. In order to allow loss guidance to exert
sufficient influence over the final image layouts when it is used in
isolation, we increase the loss-guidance strength above the level used in
iLGD (see \autoref{app:methods} for more details). All of the methods compared in this
section used the official Stable Diffusion v1.4 model~\cite{rombach2022high}
from HuggingFace. }

\paragraph{Evaluation Metrics} 
We employ a variety of metrics to measure performance along various aspects.
First, we use the T2I-Sim metric~\cite{xie2023boxdiff} to measure
text-to-image similarity between prompts and their corresponding generated
images. This metric measures the cosine similarity
between text and images in CLIP feature space~\cite{radford2021learning} to
evaluate how well the generated images reflect the semantics of
the prompt. 

We also use CLIP-IQA~\cite{wang2022exploring} to assess the quality of the
generated images. Given a pair of descriptors $\{\mathbf{y}_1,
\mathbf{y}_2\}$ which are opposite in meaning (e.g., high quality, low
quality), CLIP-IQA compares the CLIP features of these prompts with the CLIP
features of the generated image. The final score reflects how well
$\mathbf{y}_1$, as opposed to $\mathbf{y}_2$, describes the image. We
evaluate the overall quality of images using the pair $\{\text{high quality,
low quality}\}$, blurriness using $\{\text{clear, blurry}\}$, and
naturalness using $\{\text{natural, synthetic}\}$. 



To evaluate each method's faithfulness to the prescribed bounding
box, we use YOLOv4~\cite{bochkovskiy2020yolov4} to compare the predicted
bounding box over the set of generated images to the ground truth bounding
boxes, and report the average precision at $\text{IOU}=0.5$.  

Finally, we report the average contrast and saturation of generated
images. We observe that guidance-based methods often lead to high contrast
and high saturation, particularly when the guidance strength is high. 

More details about these evaluation metrics can be found in
\autoref{app:methods}.

\subsection{Comparisons}

\begin{figure*}
    \centering
    \includegraphics[width=\textwidth]{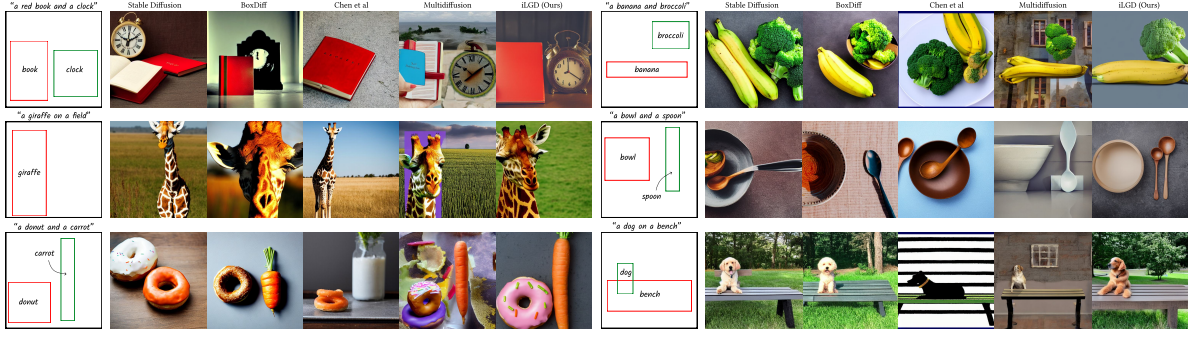}
    \caption{A comparison of iLGD against BoxDiff\revision{, Chen~\etal,
    MultiDiffusion,} and Stable Diffusion. The random seed is kept the same
    across each set of images.}
    \label{fig:ilgd_vs_boxdiff}
\end{figure*}
\begin{figure*}
    \centering
    \includegraphics[width=0.7\textwidth]{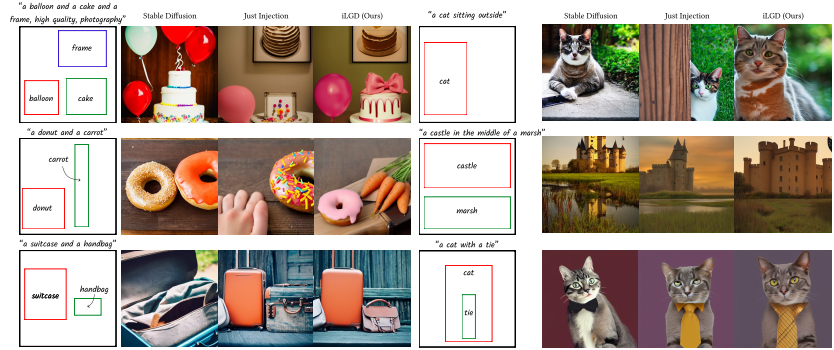}
    \caption{Images generated using either just injection or iLGD. The
    random seed is kept the same across each set of images. 
    For the prompt ``a castle in the middle of a marsh,'' we use $\nu'=0.6$ for injection and use $\eta=0.48$ when we
    additionally introduce loss guidance in iLGD.}
    \label{fig:ilgd_vs_inject}
\end{figure*}

\begin{table}[htbp]
    \centering
    \small
    \caption{Comparison of the average contrast and saturation over 200
    images for BoxDiff, \revision{Chen~\etal, MultiDiffusion, 
    Stable Diffusion (SD), with injection (SD + I) and with
    loss guidance (SD + LG), and iLGD.}}
    \footnotesize
    \begin{tabular}{l*{3}{c}}
        \toprule
                                  &    Average Contrast & Average Saturation  \\
        \midrule
        SD @ CFG 0.0              &      52.96          &  84.49  \\
        SD @ CFG 7.5              &      58.53          &  110.14 \\
        SD @ CFG 12.5             &      65.87          &  123.32 \\
        BoxDiff                   &      65.68          &  115.67 \\
        \revision{Chen \etal}     &      62.12          &  98.56  \\
        \revision{MultiDiffusion} &      48.03          &  75.01  \\
        SD + I                    &      46.07          &  105.63 \\
        SD + LG                   &      58.51          &  102.62 \\
        Ours (iLGD)               &     47.53  &  105.81 \\
        \bottomrule
    \end{tabular}
    \label{tab:contrast}
\end{table}

\begin{table}[htbp]
  \centering
  \small
  \caption{Comparison of various quality metrics for BoxDiff,
  \revision{Chen~\etal, MultiDiffusion, Stable Diffusion (SD), with
  injection (SD + I) and with loss guidance (SD + LG), and iLGD, averaged
  over 200 images.}}
    \footnotesize
    \begin{tabular}{lccccc}
        \toprule
          Method  & T2I-Sim ($\uparrow$) \hspace*{-1em}
                  & \multicolumn{3}{c}{CLIP-IQA ($\uparrow$)}
                  & \multicolumn{1}{c}{\hspace*{-1em} AP@0.5 ($\uparrow$)} \\
        \midrule
        \multicolumn{2}{c}{} & \multicolumn{1}{c}{Quality} & \multicolumn{1}{c}{Natural} & \multicolumn{1}{c}{Clear} & \\
        \midrule
          SD & 0.303 & 0.928 & \textbf{0.705} & 0.736 & -- \\
          BoxDiff & 0.305 & 0.922 & 0.613 & 0.6945 & 0.192  \\
          \revision{Chen \etal} & 0.301 & 0.936 & 0.640 & 0.814 & 0.118 \\ 
          \revision{MultiDiffusion} & 0.295 & 0.920 & 0.547 & 0.792 & \textbf{0.411} \\
          SD + I & 0.305 & 0.958  & 0.647 &  0.808 & 0.136 \\
          SD + LG \hspace*{-2em} & 0.302 & 0.932 & 0.684 & 0.737 & 0.055\\
          Ours (iLGD) \hspace*{-1em} & \textbf{0.309} & \textbf{0.961} & 0.654 & \textbf{0.817} & 0.202 \\
        \bottomrule
    \end{tabular}
  \label{tab:quality_comparison}
\end{table}


\revision{We present a comparison between our proposed method (iLGD),
BoxDiff~\cite{xie2023boxdiff}, Chen
\etal~\cite{chen2023trainingfree},
MultiDiffusion~\cite{bar2023multidiffusion}, and Stable
Diffusion~\cite{rombach2022high} in~\autoref{fig:ilgd_vs_boxdiff}.
Qualitatively, we
observe that BoxDiff and Chen~\etal both produce images with
higher contrast, and BoxDiff produces especially 
saturated images. For instance, for the prompt ``a donut and a carrot,'' both
objects in BoxDiff's image appear unnaturally bright or dark in certain
regions. 
Chen \etal produces high contrast images for the prompts ``a dog on a bench''
and ``a banana and broccoli,'' though with visibly less saturation than BoxDiff.
High contrast and saturation are even visible in some images when just
Stable Diffusion is used, e.g., in the prompts ``a balloon and a cake and a
frame\ldots'' and ``a suitcase and a handbag'' in~\autoref{fig:ilgd_vs_inject}.
}

These observations are reflected quantitatively in~\autoref{tab:contrast},
which reports the average contrast and saturation across all generated
images.  
The same high-contrast phenomenon \revision{observed in
BoxDiff and Chen \etal} is seen
to occur in Stable Diffusion when increasing the classifier-free guidance
(CFG) scale. \revision{Furthermore, high staturation can also be seen in
both BoxDiff and Stable Diffusion at high CFG scales.}
Both loss guidance and CFG appear to move the predicted latent into similar
low-probability regions
if the strength is too large, which may nonetheless be necessary to obtain
either the appropriate layout or the appropriate agreement with the
semantics of the text prompt. \revision{Neither attention injection
nor MultiDiffusion}
appear to have
this biasing effect,  and the contrast and saturation for iLGD are similar
to those of attention injection.

We note that both attention injection and iLGD produce images of
lower contrast than even Stable Diffusion without classifier-free guidance,
although visual inspection suggests that this does not manifest as any
obvious abnormality in the images. We hypothesize that injection favours
scenes that naturally contain lower contrast. When the objects appearing in
the images receive high levels of attention as a result of injection, they
tend to be clear, in-focus, and easily
discernible, and do not have either very bright reflections or very dark
shadows.

Biasing the latents using injection also clearly helps to preserve
image quality and achieve stronger layout control, particularly for layouts
which are difficult for the model to generate. For the prompt
``a red book and a clock,'' BoxDiff struggles to move
the clock into the correct position, instead generating it as a lower
quality, shadow-like figure\revision{, while Chen \etal omits the clock entirely,
and MultiDiffusion generates a chaotic scene with many artifacts.
Using iLGD, the clock and book appear
in the correct positions, and the clock} maintains its fine
details, e.g., the numbering on the face, as well as its texture. For the
prompt ``a bowl and a spoon,'' the image produced by iLGD is more faithful
to the bounding boxes than image produced by BoxDiff\revision{,
in which it is also difficult to distinguish the spoon from its shadow as dark
colours are exaggerated, as well as the image of Chen \etal, which struggles
to reposition the objects. The layout produced by MultiDiffusion is more
accurate, but the image itself is not sensible. In general, it appears that
BoxDiff performs very well, but produces images of unnaturally high contrast 
and saturation. The method of Chen \etal produces images of reasonable quality, but
struggles to move the objects
to the desired bounding boxes. MultiDiffusion appears to consistently make
objects appear in the bounding boxes, but often produces chaotic scenes or
scenes with cut-and-paste artifacts, in which the foreground images seem
glued to an incompatible background (see the prompt ``a banana and
broccoli'' for one such example).}

In~\autoref{tab:quality_comparison}, we report various metrics related to
image quality (CLIP-IQA), bounding box precision (YOLOv4) and text-to-image
similarity (T2I-Sim). \revision{While
both BoxDiff and iLGD achieve similar T2I-Sim and
AP@0.5 scores, the latter achieves better performance on all CLIP-IQA
metrics. 
While the method of Chen~\etal achieves T2I-Sim and CLIP-IQA scores that
are almost as good as those of iLGD, it achieves a much lower AP@0.5 score.
MultiDiffusion has the highest
AP@0.5 score of all methods considered, but it also has the
lowest T2I-Sim and CLIP-IQA metrics (all except for Clear), showing that
MultiDiffusion sacrifices an unacceptable amount of image quality in
exchange for layout control.
The ablation study (SD + I) using just injection shows good image quality,
but substantially worse layout control than iLGD, while the ablation study
(SD + L) using just loss guidance shows slightly worse image quality and
much worse layout control than iLGD. 
These results indicate that iLGD is able to achieve superior image quality
without sacrificing layout control, qualifying it as a meaningful
improvement over BoxDiff, Chen~\etal, and MultiDiffusion.}


We provide additional comparisons in \autoref{app:experiments}.

\subsection{Ablation Studies}

\revision{A visual comparison between iLGD and injection alone is presented in
\autoref{fig:ilgd_vs_inject}.} We observe that, while injection typically
does generate an object in
each bounding box, the object itself may be incorrect. To illustrate, for the prompt ``a balloon, a cake, and a frame, \ldots,'' using injection leads to a frame appearing where the cake should be; for ``a donut and a carrot,'' it generates a hand where the donut should be; for ``a suitcase and a handbag,'' a second suitcase is generated instead of a handbag; and for ``a cat sitting outside,'' something akin to a tree stump appears instead of a cat. In the example for ``a castle in the middle of a marsh,'' the castle does not fill up the bounding box, and for ``a cat with a tie'' the resulting image has the correct layout, but begins to appear like a cartoon cutout. When loss guidance is added using iLGD, all of these images appear correctly.
These observations are reflected in~\autoref{tab:quality_comparison}, where injection achieves a noticeably lower AP@0.5 score compared to iLGD. 

We note that the results in~\autoref{fig:ilgd_vs_inject} provide some additional evidence that injection is able to successfully bias the image according to the desired layout. For instance, revisiting the example of the prompt ``a cat sitting outside,'' injection causes a tree stump to appear in the region where the attention was enhanced. When loss guidance is also applied in each step, this stump is instead denoised into a cat. In this case, a small amount of loss guidance suffices to generate the appropriate layout, as it is augmented by the biasing effect of injection.


\section{Conclusions}

In this work, we introduce a framework which combines both attention
injection and loss guidance to produce samples conforming to a desired
layout. We show, both qualitatively and quantitatively, that our proposed
method can produce such samples with fewer visual artifacts compared to
training-free methods using loss guidance alone. Our method uses only
existing components of the diffusion model, avoiding any additional model
complexity. One of the method's limitations is that it remains somewhat
sensitive to the initial random seed. 


\appendix

\section{Diffusion Models}
\setcounter{figure}{0}
\label{app:diffusion}

\subsection{Denoising Diffusion Probabilistic Models}

Diffusion models~\cite{ho2020denoising} are characterized by two principle
algorithms. The first is the forward process, wherein the data
$\mathbf{x}_0$ is gradually corrupted by Gaussian noise until it becomes
pure noise, which we denote by $\mathbf{x}_T$. The reverse process moves in
the opposite direction, attempting to recover the data by iteratively
removing noise. The denoiser $\bm{\epsilon}_\theta(\mathbf{x}_t,t)$ is typically
a UNet~\cite{ronneberger2015unet} which accepts an image $\mathbf{x}_t$, and
predicts its noise content $\bm{\epsilon}$. Removing a fraction of this noise
yields a slightly denoised image $\mathbf{x}_{t-1}$. Repeating this
process over $T$ steps produces a noise-free image $\mathbf{x}_0$.

Operating directly on the image $\mathbf{x}_t$ in pixel-space is
computationally expensive. As an alternative, latent diffusion models have
been proposed to curtail this high cost, in which the denoising procedure is
performed in latent space, whose dimensionality is typically much lower than
pixel space. Stable Diffusion~\cite{rombach2022high} is one example of a
latent diffusion model which achieves state-of-the-art performance on
various image synthesis tasks. It leverages a powerful autoencoder to
project to and from latent space, where the standard denoising procedure is
performed. Images in latent space are typically denoted by $\mathbf{z}_t$,
and the encoder and decoder are denoted by $\mathcal{E}$ and $\mathcal{D}$,
respectively, $\mathbf{z}_t = \mathcal{E}(\mathbf{x}_t)$ and
$\mathbf{x}_t = \mathcal{D}(\mathbf{z}_t)$.

During training, samples from the true data distribution $q(\mathbf{x}_0)$
are corrupted via the forward process. By training a diffusion model to
learn a reverse process in which it iteratively reconstructs these noisy
samples into noise-free samples, it is possible to generate images from pure
noise at inference time. This corresponds to sampling from an approximation
$p_\theta(\mathbf{x}_0)$ to the data
distribution, $q(\mathbf{x}_0)$. This generation process can be guided by
introducing an additional input vector $\mathbf{y}$, which is often a text
prompt. In this case, the model produces samples from an
approximation $p_\theta(\mathbf{x}_0|\mathbf{y})$ to the conditional
distribution $q(\mathbf{x}_0|\mathbf{y})$. 

In denoising diffusion probabilistic models (DDPM)~\cite{ho2020denoising}, the
forward process is characterized by the Markov chain $q(\bx_t|\bx_{t-1})
\sim \mathcal{N}(\bx_t; \sqrt{1-\beta_t} \bx_{t-1},\beta_t \mathbf{I})$, for
some noise schedule $\beta_t$. In this case, $q(\bx_t|\bx_0) \sim
\mathcal{N}(\bx_t;\sqrt{\bar \alpha_t} \bx_0, (1-\bar \alpha_t)
\mathbf{I})$, where $\bar \alpha_t = \prod_{s=1}^t \alpha_s$ and $\alpha_t =
1-\beta_t$. The reverse process is typically modeled by a learned Markov
chain $p_\theta(\bx_{t-1}|\bx_t) \sim \mathcal{N}(\bx_{t-1};
\mu_\theta(\bx_t,t),\sigma_t^2 \mathbf{I})$, \revision{where $\sigma_t$ is
an untrained time dependent constant, usually with $\tilde \beta_t \le \sigma_t^2
\le \beta_t$ and $\tilde \beta_t = \beta_t (1-\bar \alpha_{t-1})/(1-\bar
\alpha_t)$, or with $\sigma_t$ simply chosen equal to $\sqrt{\beta_t}$.}

It is not efficient to optimize the log-likehood $\mathbb{E}[-\log
p_\theta(x_0)]$ directly, since computing $p_\theta(\bx_0)$ requires
marginalizing over $\bx_{1:T}$. Instead, one can use importance sampling
to write
\begin{align}
p_\theta(\bx_0) = \mathbb{E}_{q(\bx_{1:T}|\bx_0)} \left[ 
\frac{p_\theta(\bx_{0:T})}{q(\bx_{1:T}|\bx_0)} \right].
\end{align}
Then, by Jensen's inequality,
\begin{align}
-\log p_\theta(\bx_0) \le \mathbb{E}_{q(\bx_{1:T}|\bx_0)} \left[ 
-\log \frac{p_\theta(\bx_{0:T})}{q(\bx_{1:T}|\bx_0)} \right].
\end{align}
The right hand side is the usual evidence lower bound (ELBO), which is
minimized instead. Ho \etal~\cite{ho2020denoising} show that
minimizing the ELBO is equivalent to minimizing
\begin{align}
\mathbb{E}_t [ \lambda(t)
    \mathbb{E}_{q(\mathbf{x}_0),\epsilon}
    [\lVert \epsilon_\theta(\mathbf{x}_t, t) - \epsilon \rVert^2_2 ]],
      \label{eq:errormin}
\end{align}
for some \revision{positive function} $\lambda(t)$, where $\epsilon \sim
\mathcal{N}(\mathbf{0},\mathbf{I})$, $\bx_t(\bx_0,\epsilon) =
\sqrt{\bar\alpha_t} \bx_0 + \sqrt{1-\bar \alpha_t} \epsilon$, and
\begin{align}
\mu_\theta(\bx_t,t) = \frac{1}{\sqrt{\alpha_t}} \Bigl( \bx_t -
\frac{\beta_t}{\sqrt{1-\bar\alpha_t}} \epsilon_\theta(\bx_t,t) \Bigr).
\end{align}

\subsection{Score Matching}

Since $q(\bx_t|\bx_0)$ is a normal distribution, we know that
\begin{align}
\nabla_{\bx_t} \log q(\bx_t | \bx_0) = -\frac{\epsilon}{\sqrt{1-\bar
\alpha_t}}.
\end{align}
Thus, minimizing~\eqref{eq:errormin} is equivalent to minimizing
\begin{align}
\mathbb{E}_t [ \lambda(t)
    \mathbb{E}_{q(\mathbf{x}_0)} \mathbb{E}_{q(\bx_t|\bx_0)}
    [\lVert \mathbf{s}_\theta(\mathbf{x}_t, t) - \nabla_{\bx_t} \log q(\bx_t|\bx_0) 
    \rVert^2_2 ]],
\end{align}
for some \revision{positive function} $\lambda(t)$, where
\begin{align}
\mathbf{s_\theta}(\mathbf{x}_t, t) :=
    -\frac{\epsilon_\theta(\mathbf{x}_t, t)}{\sqrt{1-\Bar{\alpha}_t}}.
\end{align}
It is known that this loss is minimized when
$\mathbf{s_\theta}(\mathbf{x}_t, t) = \nabla_{\bx_t} \log q(\bx_t)$~\cite{vincent2011connection},
so given enough parameters, $\mathbf{s_\theta}(\mathbf{x}_t, t)$ will
converge to $\nabla_{\bx_t} \log q(\bx_t)$ almost everywhere. 
Given an approximation to the score
function, it is possible to sample from $p_\theta(\bx_0)$ using annealed
Langevin dynamics~\cite{song2020generative}.

\revision{
Letting the forward process posterior mean $\tilde \mu_t$ be defined by
$q(\bx_{t-1}| \bx_t,\bx_0) \sim \mathcal{N}(\bx_{t-1}; \tilde
\mu_t(\bx_t,\bx_0),\tilde \beta_t \mathbf{I})$, we have that
\begin{align}
\tilde \mu_t(\bx_t,\bx_0) := \frac{\sqrt{\bar \alpha_{t-1}} \beta_t}{1-\bar
\alpha_t} \bx_0 + \frac{\sqrt{\alpha_t}(1-\bar \alpha_{t-1})}{1-\bar
\alpha_t} \bx_t
\end{align}
(see~\cite{ho2020denoising}).  With this, the mean $\mu_\theta(\bx_t,t)$ of
the reverse process can be understood as
\begin{align}
\mu_\theta(\bx_t,t) = \tilde \mu_t(\bx_t, D_\theta(\bx_t,t)),
\end{align}
where
\begin{align}
D_\theta(\bx_t,t) := \frac{1}{\sqrt{\bar \alpha_t}}(\bx_t + (1-\bar \alpha_t)
s_\theta(\bx_t,t))
\end{align}
is an approximation to Tweedie's formula
\begin{align}
\begin{split}
&\frac{1}{\sqrt{\bar \alpha_t}}(\bx_t + (1-\bar \alpha_t) \nabla_{\bx_t} \log
q(\bx_t))  \\
&\hspace*{3em}= \frac{1}{\sqrt{\bar \alpha_t}} \mathbb{E}_q[\sqrt{\bar \alpha_t}
\bx_0|\bx_t] = \mathbb{E}_q[\bx_0|\bx_t]
\end{split}
\end{align}
(see~\cite{efron2011tweedie}). }

\subsection{Stochastic Differential Equations}

Song \etal\cite{song2020score} showed that the forward 
process of DDPM can viewed as a discretization of the stochastic differential
equation (SDE)
\begin{align}
d\bx = -\frac{1}{2}\beta(t)\bx\, dt + \sqrt{\beta(t)}\, d\mathbf{w},
\end{align}
where $\mathbf{w}$ denotes the Wiener process. There, the authors point out
that any SDE of the form $d\bx = \mathbf{f}(\bx,t) + g(t)d\mathbf{w}$, where
$\bx_0 \sim p_0(\bx_0)$ can be reversed by the SDE $d\bx =
(\mathbf{f}(\bx,t) - g(t)^2 \nabla_\bx \log p_t(\bx))\, dt +g(t)\,
d\mathbf{\bar w}$, where $\mathbf{\bar w}$ is the standard Wiener process in
the reverse time direction, and where $\bx_T \sim p_T(\bx_T)$.  Furthermore,
each SDEs admits a family of related SDEs that share the same marginal
distributions $p_t(\bx_t)$. One of these SDEs is purely deterministic, and is
known as the probability flow ordinary differential equation (ODE).

If the score function $\mathbf{s}_\theta(\bx_t,t)$ is available, then it
is possible to sample from $p_\theta(\bx_0)$ by solving the probabily
flow ODE, starting with samples  from $p_\theta(\bx_T)$. This results in a 
deterministic mapping from noises images $\bx_T$ to clean images $\bx_0$.
This sampling process can be performed quickly with the aid of ODE
solvers~\cite{lu2022dpm}.

\revision{
\subsection{Classifier-Free Guidance}
}

In order to generate images following a user-supplied text prompt, the
denoiser $\bm{\epsilon}_\theta(\mathbf{z}_t,t,\mathbf{y})$ of a latent diffusion model
is trained with an additional input given by a sequence of token embeddings
$\mathbf{y}=\{\mathbf{y}_1,\mathbf{y}_2,\ldots,\mathbf{y}_k\}$. A single
denoiser, usually a UNet, is trained over a variety of text prompts, and the
token embeddings influence the denoiser by a cross-attention mechanism in
both the contractive and expansive layers. Ho and Salimans~\cite{ho2021classifierfree}
found that, rather than sampling images using the conditional denoiser
alone, better results can be obtained by taking a combination of conditional
and unconditional noise estimates,
\begin{align}
\bm{\tilde\epsilon}_\theta(\mathbf{z}_t,t,\mathbf{y})
=(1+w)\bm{\epsilon}_\theta(\mathbf{z}_t,t,\mathbf{y})
-w\bm{\epsilon}_\theta(\mathbf{z}_t,t,\{\}),
\end{align}
where $w$ represents the intensity of the additive term
$\bm{\epsilon}_\theta(\mathbf{z}_t,t,\mathbf{y})-
\bm{\epsilon}_\theta(\mathbf{z}_t,t,\{\})$.  For $-1\le w\le 0$, this noise
prediction can be viewed as an approximation to ($-\sigma_t$ times) the
score function of the marginal distribution $\tilde
p_\theta(\mathbf{z}_t|\mathbf{y}) \propto p_\theta(\mathbf{z}_t|\mathbf{y})^{1+w}
p_\theta(\mathbf{z}_t|\{\})^{-w}$.  In classifier-free guidance (CFG), 
$w\gg 0$, which does not have a simple interpretation in terms
of the marginal distributions of the new denoising process.

\section{Detailed Methods}
\setcounter{figure}{0}
  \label{app:methods}

\paragraph{Implementation Details}

We implement our method on the official Stable Diffusion v1.4
model~\cite{rombach2022high} from HuggingFace. All images are generated
using 50 denoising steps and a classifier-free guidance scale of 7.5, unless
otherwise noted. We use the noise scheduler \texttt{LMSDiscreteScheduler}
\cite{karras2022elucidating} provided by HuggingFace. Experiments are
conducted on an NVIDIA TESLA V100 GPU. 

We perform attention injection over all attention maps. When performing
injection, we resize the mask $m$ to the appropriate resolution, depending
on which layer of the UNet the attention maps are taken from. For loss guidance,
we again use all of the model's attention maps, but resize them $16\times
16$ resolution, and compute the mean of each map over all pixels.  We apply
the softmax function over these means to obtain a weight vector
$\mathbf{w}$, where each entry $w_j$ is the scalar weight associated with
the $j$-th resized attention map.  Finally, we obtain the attention map $A_t$
by taking a weighted average over all resized attention maps at time $t$,
using the appropriate weight $w_j$ for each map. 

When attempting to control the layout of a generated image, we find that
skipping the first step, so that it remains a standard denoising step, leads
to better results. We do this for all experiments conducted in this paper
which use either injection or loss guidance or both. \revision{In iLGD, we}
use $\eta=0.48$, $\nu'=0.75$, $t_{\text{loss}}=25$, and
$t_{\text{inject}}=10$, unless otherwise noted.
\revision{In our ablation experiments, we keep the injection strength 
at $\nu'=0.75$ when
performing just attention injection. When performing just loss guidance, we
increase the loss-guidance strength to $\eta=1$, in order to make loss
guidance alone exert sufficient influence over the final image layouts.}

In our comparisons with BoxDiff, we maintain the default parameters the
authors provide in their implementation. We start with $\alpha_T=20$, which
decays linearly to $\alpha_0=10$, and perform guidance for 25 iterations out
of a total of 50 denoising steps.

\revision{In our comparisons with the method of Chen \etal, we also maintain
the default parameters the authors provide in their implementation,
setting the loss scale factor to $\eta=30$.}
  
\paragraph{Evaluation with YOLOv4}
In this section, we describe in detail how we obtain the AP@50 scores in Table 2. In classical object detection, a model is trained to detect and localize objects of certain classes in an image, typically by predicting a bounding box which fully encloses the object. The accuracy of the model's predicted bounding box, $B_p$, is evaluated by comparison to the  corresponding ground truth bounding box, $B_{gt}$. More specifically, we compute the intersection over union (IOU) over the pair of bounding boxes:

\begin{equation}
    \label{eq:iou}
    \text{IOU} = \frac{\text{area}(B_p \cap B_{gt})}{\text{area}(B_p\cup B_{gt})}.
\end{equation}

The IOU is then compared to a threshold $t$, such that, if $\text{IOU} \geq t$, then the detection is classified as correct. If not, then the detection is classified as incorrect. In our case, we follow Li \etal~\cite{li2021image} and treat the object detection model as an oracle, where we assume that it provides the bounding boxes of objects in a given image with perfect accuracy. In particular, we first define a layout through a set of ground truth bounding boxes, describing the desired positions of each object. We then generate an image according to this layout, and subsequently apply the object detection model to the generated image to obtain a set of predicted bounding boxes. Finally, to evaluate how similar the layout of the generated image is to the desired layout, we compare each predicted bounding box, $B_p$, to the corresponding ground truth bounding box, $B_{gt}$,  by computing their IOU. We use a IOU threshold of 0.5. 

To calculate the average precision, we first need to compute the number of true positives (TP), false positives (FP), and false negatives (FN). We count a false negative when no detection is made on the image, even though a ground truth object exists, or when the detected class is not among the ground truth classes. We also count a false negative as well as a false positive when the correct detection is made, but $\text{IOU} < 0.5$, and a true positive when $\text{IOU} \geq 0.5$. Using these quantities, we compute the precision $P$ and recall $R$ as:

\begin{equation}
    \label{eq:precision}
    P = \frac{TP}{TP + FP},
\end{equation}

\begin{equation}
    \label{eq:recall}
    R = \frac{TP}{TP + FN}.
\end{equation}

We repeat this for classifier confidence thresholds of 0.15 to 0.95, in steps of 0.05, so that we end up with 17 values for precision and recall, respectively. We then construct a precision-recall curve, and compute the average precision using 11-point interpolation~\cite{inproceedings}:

\begin{equation}
    \label{eq:AP11}
    \text{AP}_{11} = \frac{1}{11}  \sum_{R\in \{0, 0.1, ..., 0.9, 1\}} \!\!\!\!\!\!\!\! P_{\text{interp}}(R),
\end{equation}

\noindent
where

\begin{equation}
    \label{eq:interp_precision}
    P_{\text{interp}}(R) = \max_{\tilde{R}\geq R} P(\tilde{R}).
\end{equation}

\paragraph{Image Quality Assessment} Wang \etal\cite{wang2022exploring} suggest using the pair \{good photo, bad photo\} instead of \{high quality, low quality\} to measure quality, as they find that it corresponds better to human preferences. However, we choose the latter to remain agnostic to the image's style, as we believe the former carries with it a stylistic bias, due to the word ``photo.''

\paragraph{Contrast Calculation} We calculate the RMS contrast by using OpenCV's \texttt{.std()} method on a greyscale image. 

\paragraph{Saturation Calculation} We calculate the saturation by working in HSV space and using OpenCV's \texttt{.mean()} method on the image's saturation channel.

\section{Additional Experiments}
\setcounter{figure}{0}
  \label{app:experiments}

\begin{figure*}
    \centering
    \includegraphics[width=\textwidth]{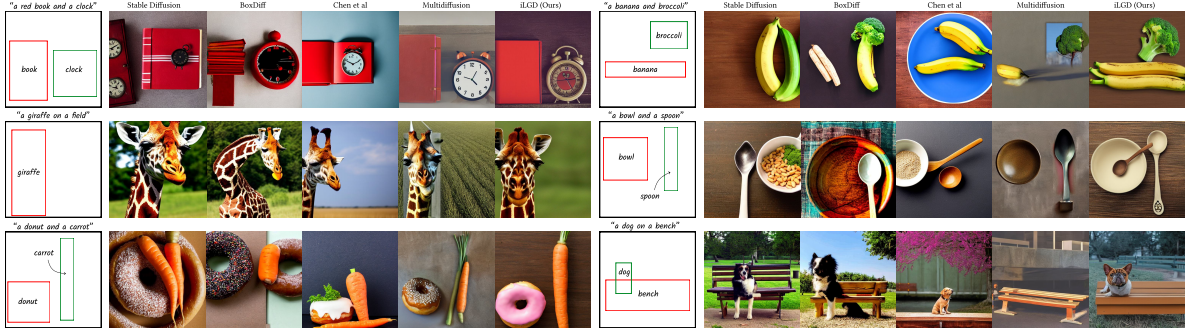}
    \caption{A comparison of iLGD against BoxDiff\revision{, Chen~\etal,
    MultiDiffusion,} and Stable Diffusion, using the
    same prompts as Figure~3 but different random seeds, with the seed kept
    the same across each set of images.}
    \label{fig:ilgd_vs_boxdiff_appdx}
\end{figure*}

We provide two additional sets of comparisons between our proposed method
(iLGD), \revision{BoxDiff~\cite{xie2023boxdiff}, Chen
\etal~\cite{chen2023trainingfree},
MultiDiffusion~\cite{bar2023multidiffusion}, and Stable
Diffusion~\cite{rombach2022high}}. In
\autoref{fig:ilgd_vs_boxdiff_appdx}, we compare the three methods using same
prompts and bounding boxes as in Figure~3, but using a different random seed
for each set of images. In \autoref{fig:ilgd_vs_boxdiff_appdx_2}, we compare
the methods using an entirely new set of prompts and bounding boxes. 

\begin{figure*}
    \centering
    \includegraphics[width=\textwidth]{images/ilgd_vs_boxdiff_appendix_2.pdf}
    \caption{A comparison of iLGD against BoxDiff\revision{, Chen~\etal,
    MultiDiffusion,} and Stable Diffusion. The random
    seed kept the same across each set of images.}
    \label{fig:ilgd_vs_boxdiff_appdx_2}
\end{figure*}

\newpage

{\small
\bibliographystyle{ACM-Reference-Format}
\bibliography{main}
}

\end{document}